\documentclass[10pt,twocolumn,letterpaper]{article}

\usepackage[pagenumbers]{cvpr} 

\usepackage{graphicx}
\usepackage{amsmath}
\usepackage{amssymb}
\usepackage{booktabs}

\usepackage{colortbl}
\usepackage{graphicx}

\usepackage{blindtext}

\makeatletter
\@namedef{ver@everyshi.sty}{}
\makeatother
\usepackage{tikz}

\usepackage[precision=2, unit=mm]{lengthconvert}
\usetikzlibrary{patterns}

\usepackage[pagebackref,breaklinks,colorlinks]{hyperref}
\usepackage{xcolor}
\usepackage{siunitx}
\usepackage{blindtext}
\usepackage[normalem]{ulem}

\newcommand{\anonymouslink}[1]{anonymous link}

\usepackage[capitalize]{cleveref}
\crefname{section}{Sec.}{Secs.}
\Crefname{section}{Section}{Sections}
\Crefname{table}{Table}{Tables}
\crefname{table}{Tab.}{Tabs.}

\usepackage{enumitem}

\usepackage{algorithm}
\usepackage{algpseudocode}

\usepackage{url}

\pagecolor{white}

\begin{document}

\title{The Stable Artist\\ Interacting with Concepts in Diffusion Latent Space}
\title{The Stable Artist:\\ Steering Semantics in Diffusion Latent Space}

\author{
Manuel Brack$^{1}$
\and
Patrick Schramowski$^{1,3,4,5}$
\and
Felix Friedrich$^{1,3}$ 
\and
Dominik Hintersdorf$^{1}$
\and
Kristian Kersting$^{1,2,3,4}$
\and
$^{1}$Computer Science Department, TU Darmstadt\\
$^{2}$Centre for Cognitive Science, TU Darmstadt, 
$^{3}$Hessian Center for AI (hessian.AI)\\
$^{4}$German Research Center for Artificial Intelligence (DFKI),  $^{5}$LAION\\
{\tt\small \{lastname\}@cs.tu-darmstadt.de}
}

\twocolumn[{
\maketitle
\begin{center}
    \captionsetup{type=figure}
    \includegraphics[width=.85\linewidth]{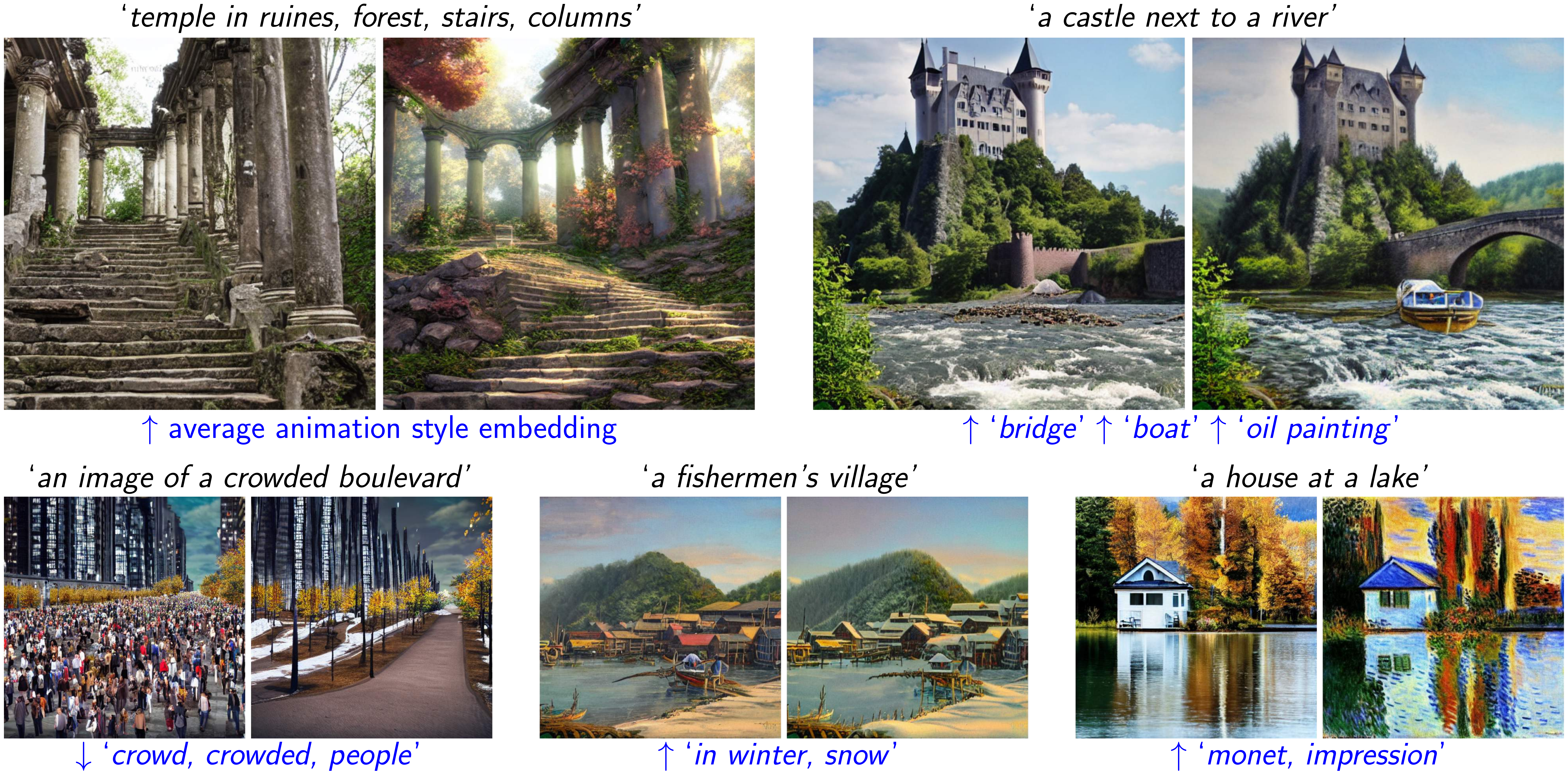}
    \captionof{figure}{The Stable Artist eases the generation of images via Stable Diffusion by (iterative) guidance. Original image (left images) generated using the prompt on top of image pair.  Guidance prompt (bottom of image pair) and result (right). (Best viewed in color)  }
    \label{fig:examples}
\end{center}
}]

\begin{abstract}
Large-scale, text-conditioned generative diffusion models have recently drawn a lot of interest for their astonishing ability to produce high-fidelity images from text only. 
However, generating high-quality images in a single shot is nearly impossible and typically requires several small changes to the input prompt. 
Unfortunately, small changes to the input prompt often result in completely different images being generated, 
leaving the artist with little control.

We present the Stable Artist to enable control by allowing the artist 
to steer the diffusion process along a variable number of semantic directions.
This semantic guidance (\textsc{Sega}) allows for subtle edits to images, changes in composition and style, as well as optimization of the overall artistic conception. Furthermore, \textsc{Sega} enables probing latent spaces to gain insights into the representation of concepts learned 
by the model, even complex ones such as `carbon emission'. We demonstrate the Stable Artist on several tasks, showcasing high-quality image editing and composition. 
\end{abstract}

\noindent

 \section{Introduction}

Generative AI methods are improving rapidly, and using text-to-image diffusion models (DM)~\cite{saharia2022photorealistic, ramesh2022hierarchical, rombach2022High}
makes it now possible to generate text and images simply based on text input, producing 
impressive results on generative image tasks.
Unfortunately, however, unraveling the concepts they learn during training and understanding how to influence what they actually output remains an open question. Users need deep knowledge of style-specific, long, and obfuscated prompts that usually eludes non-experts. Hence, high-quality images are rarely the result of the initially generated output, making a one-shot text-to-image generation infeasible.
The contrary is the case; text-guided image generation is a highly iterative process that requires interaction between the model and its user. 
Consequently, the human user is likely to generate many images with slight changes to the text prompt and other parameters in order to achieve the envisioned outcome. In this `artist-in-the-loop' setting, fine-grained control over the generated output and its elements is imperative but hardly feasible through current methods. Small changes in the phrasing of the prompt text may lead to the generation of entirely different images.  

The required amount of control is generally only possible through providing image masks in combination with an edit instruction of the masked area. This is inherently limited, as it discards important structural information and the global composition of the image. Furthermore, some tasks like texture or style changes are not achievable with inpainting techniques. Other approaches like Prompt-to-Prompt (P2P) \cite{hertz2022prompt} rely on a form of soft, implicit masking by interacting with the attention masks of the input prompt. P2P utilizes the changes in attention maps between the original prompt and an edited prompt to target the relevant regions of the image. However, the granularity of control remains limited to the rather core-grained dimensions of the attention mask, and these approaches are inherently restricted to one editing operation at a time. 
On the other hand, Composable Diffusion \cite{liu2022Compositional} does enable conditioning on multiple concepts but only provides control over the initial image composition and does not support more subtle changes. 

We present the Stable Artist, an iterative approach for guiding a generated image toward the desired output. The Stable Artist can change aspects of the initial image using Semantic Guidance (\textsc{Sega}) that provides fine-grained control of the image generation process by leveraging sophisticated operations in the model`s latent space. This enables subtle edits to images, changes in composition and style, as well as optimization of the overall artistic conception. Furthermore, \textsc{Sega} allows for probing the latent space of diffusion models to gain insights into how abstract concepts are represented by the model and how their interpretation reflects on the generated image. \textsc{Sega} also facilitates advanced arithmetics between concepts that were previously only observed for natural language embeddings \cite{mikolov-etal-2013-linguistic,levy-goldberg-2014-linguistic, agirre-etal-2012-semeval}. 

The Stable Artist supports editing with multiple concepts simultaneously while providing full control over the extent of changes to the image as well as the strength of each applied concept. All while using no masks---be it explicit or based on attention---and without any fine-tuning. 


\section{The Stable Artist}
\begin{algorithm}[t!]
\small
\caption{\underline{Se}mantic \underline{G}uid\underline{a}nce (\textsc{Sega})}\label{alg:sega}
\begin{algorithmic}
\Require model weights $\theta$, text condition $text_p$, edit texts \textbf{List}(${text_e}$) and diffusion steps $T$
\Ensure $s_m \in [0,1]$, $\nu_{t=0}=0$, $\beta_m \in [0,1)$, $\lambda^i \in [-1,1]$, $s_e^i \in [0,5000]$, $\delta \in [0,20]$, $t = 0$
\State $\text{DM} \gets \text{init-diffusion-model}(\theta)$
\State $c_p \gets \text{DM}.\text{encode}(text_p)$
\State $\textbf{List}(c_e) \gets \text{DM}.\text{encode}(\textbf{List}(text_e))$
\State $latents \gets \text{DM}.\text{sample}(seed)$
\While{$t \neq T$}
    \State $n_\emptyset, n_p \gets \text{DM}.\text{predict-noise}(latents, c_p)$
    \State $\textbf{List}(n_e) \gets \text{DM}.\text{predict-noise}(\textbf{List}(c_e))$
    \State $\mu_t \gets \mathbf{0}$ \Comment{\cref{eq:threshold}\phantom{0}}
    \ForAll{$n_e$ in $\textbf{List}(n_e)$}
        
        \State $\phi_t^i \gets s_e^i *(n_p - n_e)^{-1}$\Comment{\cref{eq:safety_value}\phantom{0}}
        \If{positive guidance}
            \State $\mu_t^i \gets \text{where}(n_p - n_e > \lambda, \text{max}(1,|\phi_t|))$ \Comment{\cref{eq:threshold}\phantom{0}}
            \State $\gamma_t^i \gets \mu_t^i*(n_\emptyset-n_e) $ \Comment{\cref{eq:sem_guidance,eq:guidance_direction}\phantom{0}}
            \Else
                \State $\mu_t^i \gets \text{where}(n_p - n_e < \lambda, \text{max}(1,|\phi_t|))$ \Comment{\cref{eq:threshold}\phantom{0}}
                \State $\gamma_t^i \gets \mu_t^i*(n_e-n_\emptyset) $ \Comment{\cref{eq:sem_guidance,eq:guidance_direction}\phantom{0}}
        \EndIf
    \EndFor
    \State $\gamma_t \gets \sum\nolimits_{i \in I} g_i * \gamma_t^i $ \Comment{\cref{eq:mult_sumup}}
    \State $\gamma_t \gets \gamma_t + s_m*\nu_t$ \Comment{\cref{eq:safety_guidance}\phantom{0}}
    
    \State $\nu_{t+1} \gets \beta_m * \nu_t (1 - \beta_m)*\gamma_t$ \Comment{\cref{eq:safety_momentum}\phantom{0}}
    \If{$t \geq \delta$}
        \State $pred \gets s_g*(n_p - n_\emptyset - \gamma_t)$ \Comment{\cref{eq:final_noise_pred}\phantom{0}}
    \Else
        \State $pred \gets s_g*(n_p - n_\emptyset)$ \Comment{\cref{eq:classifier_free}\phantom{0}}
    \EndIf
    \State $latents \gets \text{DM}.\text{update-latents}(pred, latents)$
    \State $t \gets t + 1$
\EndWhile
\State $image \gets \text{DM}.\text{decode}(latents)$
\end{algorithmic}
\end{algorithm}

\begin{figure*}
    \centering
    \begin{subfigure}[t]{0.48\textwidth}
         \centering
         \includegraphics[width=\linewidth]{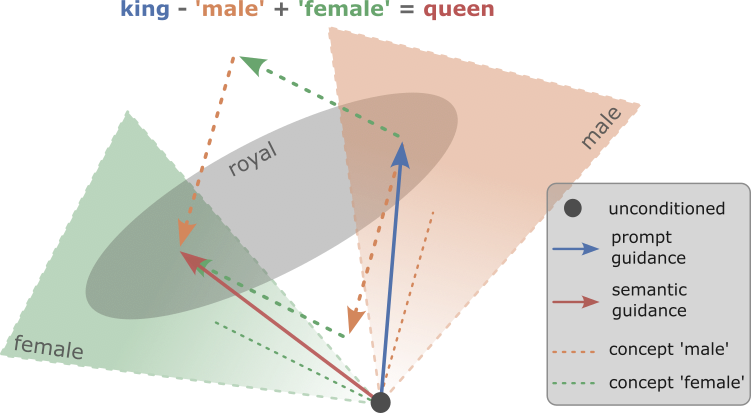}
         \caption{A (latent) diffusion process inherently organizes concepts and learns implicitly relationships between them, although there is no supervision.}
         \label{fig:arithmetics_graph}
     \end{subfigure}
     \hfill
     \begin{subfigure}[t]{0.48\textwidth}
         \centering
         \includegraphics[width=.9\linewidth]{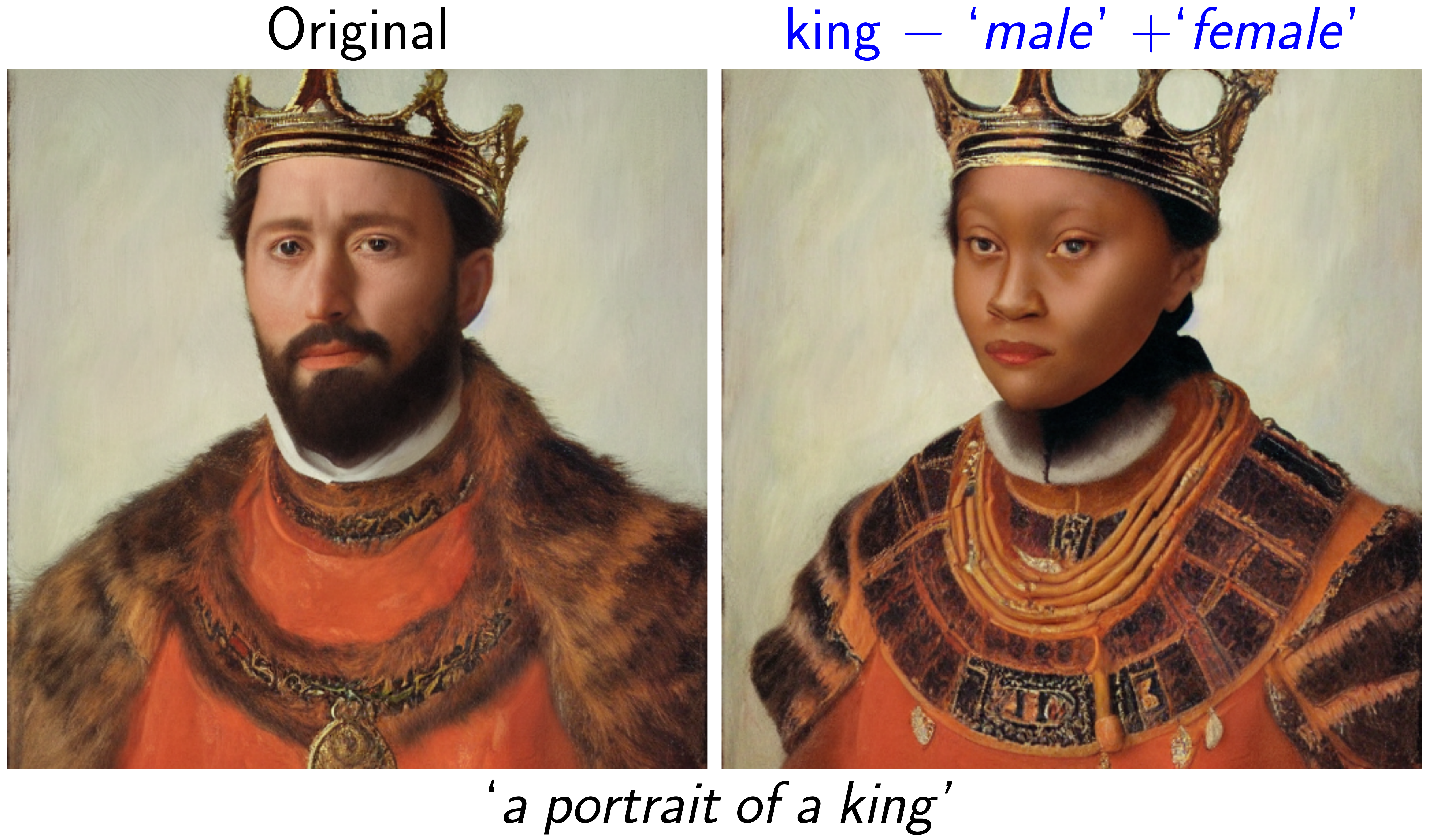}
                 \caption{Guidance arithmetic:  Guiding the image `a portrait of a king' (left) using `king'$-$‘male’$+$‘female' results in an images of a `queen' (right).}
         \label{fig:arithmetics_example}
     \end{subfigure}
    \caption{Semantic guidance (\textsc{SEGA})  applied to the image `a portrait of a king' using `king'$-$‘male’$+$‘female'. (Best viewed in color)}
    \label{fig:arithmetics}
\end{figure*}

Let us now devise the Stable Artist\footnote{In this context, `Stable' is not only a nod to Stable Diffusion that builds the base of our implementation. Additionally, the Stable Artist behaves stable with respect to its control over the generated image.}, a semantic image editing technique for latent diffusion models. Roughly speaking, it generalizes the generative diffusion process by combining text conditioning through classifier-free guidance with editing concepts targeting dedicated parts of the image. In doing so, it substantially extends the combination of latents introduced in composable diffusion \cite{liu2022Compositional} and is a general-purpose advancement of Safe Latent Diffusion (SLD) \cite{schramowski2022safe}; while SLD suppresses one inappropriate direction during generation, the Stable Artist features a variable number of directions that can be
suppressed or enforced. 

\subsection{Guided Diffusion}
The first step towards the Stable Artist is guided diffusion. Specifically, diffusion models iteratively denoise a Gaussian distributed variable to produce samples of a learned data distribution. For text-to-image generation, the model is conditioned on a text prompt $p$ and guided towards an image, faithful to that prompt. The training objective of a diffusion model $\hat{x}_\theta$, can be written as
\begin{equation}
    \mathbb{E}_{\mathbf{x,c}_p\mathbf{,\epsilon},t}\left[w_t||\mathbf{\hat{x}}_\theta(\alpha_t\mathbf{x} + \omega_t\mathbf{\epsilon},\mathbf{c}_p) - \mathbf{x}||^2_2 \right]
\end{equation}
where $(\mathbf{x,c}_p)$ is conditioned on text prompt $p$, $t$ is drawn from a uniform distribution $t\sim\mathcal{U}([0,1])$, $\epsilon$ sampled from a Gaussian $\mathbf{\epsilon}\sim\mathcal{N}(0,\mathbf{I})$, and $w_t, \omega_t, \alpha_t$ influence image fidelity depending on $t$. 
Consequently, the DM is trained to denoise $\mathbf{z}_t := \mathbf{x}+\mathbf{\epsilon}$ to yield $\mathbf{x}$ with the squared error as a loss. At inference, the DM is sampled using the model's prediction of $\mathbf{x}=(\mathbf{z}_t - \mathbf{\Bar{\epsilon_\theta}})$, with ${\Bar{\epsilon_\theta}}$ as described below. 
%
%

%

Classifier-free guidance \cite{ho2022classifier} is a conditioning method using a purely generative diffusion model, eliminating the need for an additional pre-trained classifier. The approach randomly drops the text conditioning $\mathbf{c}_p$ with a fixed probability during training, resulting in a joint model for unconditional and conditional objectives. 
During inference the score estimates for the $\mathbf{x}$-prediction are adjusted so that: 
\begin{equation}\label{eq:classifier_free}
    \mathbf{\Tilde{\epsilon}}_\theta(\mathbf{z}_t, \mathbf{c}_p) := \mathbf{\epsilon}_\theta(\mathbf{z}_t) + s_g (\mathbf{\epsilon}_\theta(\mathbf{z}_t, \mathbf{c}_p) - \mathbf{\epsilon}_\theta(\mathbf{z}_t))
\end{equation}
with guidance scale $s_g$ and $\epsilon_\theta$ defining the noise estimate with parameters $\theta$. Intuitively, the unconditioned $\epsilon$-prediction is pushed in the direction of the conditioned one, with the $s_g$ determining the extent of the adjustment.

\subsection{Semantic Guidance}
Now, the main idea is to  
influence the diffusion process along several directions. To achieve this, \textsc{Sega} substantially extends the principles introduced in classifier-free guidance.
The idea is to use 
multiple editing prompts $e_i$ targeting arbitrary concepts of the generated image, in addition to the text prompt $p$. 

\paragraph{One Direction.}
To introduce semantic guidance, let us start off with a single direction, i.e., editing prompt.
Specifically, we use three $\epsilon$-predictions with the goal of moving the unconditioned score estimate $\mathbf{\epsilon}_\theta(\mathbf{z}_t)$ towards the prompt conditioned estimate $\mathbf{\epsilon}_\theta(\mathbf{z}_t, \mathbf{c}_p)$ and simultaneously away or also towards the concept conditioned estimate  $\mathbf{\epsilon}_\theta(\mathbf{z}_t, \mathbf{c}_e)$, depending on the editing direction. Formally, we compute 
\begin{align}
\label{eq:final_noise_pred}
\mathbf{\Bar{\epsilon}}_\theta(&\mathbf{z}_t, \mathbf{c}_p, \mathbf{c}_e)= \nonumber \\
       &\mathbf{\epsilon}_\theta(\mathbf{z}_t) + s_g \big(\mathbf{\epsilon}_\theta(\mathbf{z}_t, \mathbf{c}_p) - \mathbf{\epsilon}_\theta(\mathbf{z}_t) - \gamma(\mathbf{z}_t, \mathbf{c}_p, \mathbf{c}_e)\big) 
\end{align}
with the semantic guidance term $\gamma$
\begin{equation}
    \gamma(\mathbf{z}_t, \mathbf{c}_p, \mathbf{c}_e) = \mu(\mathbf{c}_p, \mathbf{c}_e; s_e, \lambda) \psi(\mathbf{z}_t, \mathbf{c}_p, \mathbf{c}_e)
    \label{eq:sem_guidance}
\end{equation}
where $\mu$ applies an editing guidance scale $s_e$ element-wise, and $\psi$ depends on the editing direction:
\begin{align}
    \psi(\mathbf{z}_t, &\mathbf{c}_p, \mathbf{c}_e)=\nonumber\\ 
    &\begin{cases}
 \mathbf{\epsilon}_\theta(\mathbf{z}_t) - \mathbf{\epsilon}_\theta(\mathbf{z}_t, \mathbf{c}_e)  &  \text{if pos. guidance} \\
   -(\mathbf{\epsilon}_\theta(\mathbf{z}_t) - \mathbf{\epsilon}_\theta(\mathbf{z}_t, \mathbf{c}_e)) &  \text{if neg. guidance}
   \label{eq:guidance_direction}
\end{cases}
\end{align}
Consequently, changing the guidance direction is reflected by the direction of the vector between $\mathbf{\epsilon}_\theta(\mathbf{z}_t, \mathbf{c}_e)$ and $\mathbf{\epsilon}_\theta(\mathbf{z}_t)$.
\begin{figure*}
\centering
     \includegraphics[width=.9\linewidth]{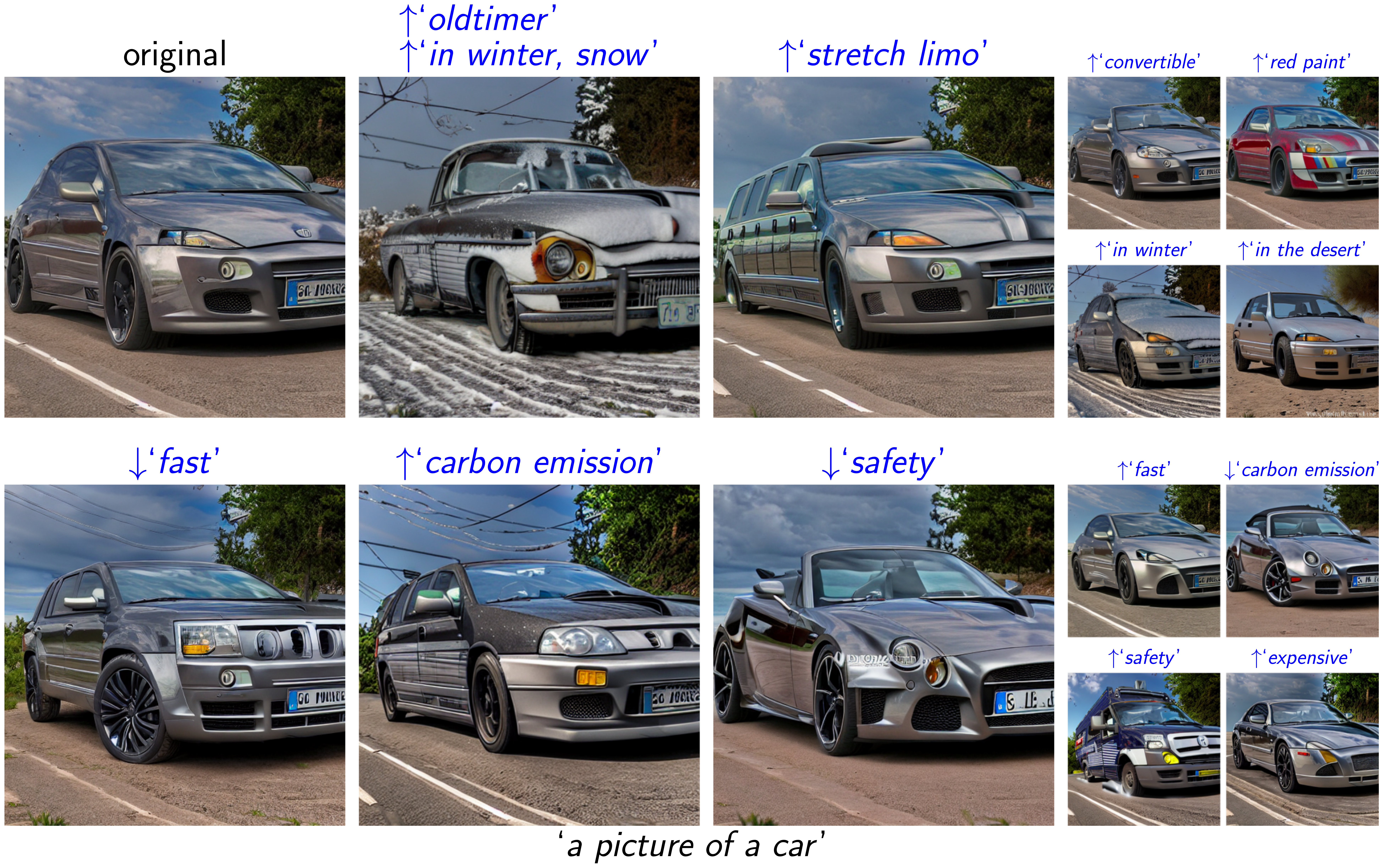}
     \caption{Image editing, performed using semantic guiding of the Stable Artist. All images generated from the same initial noise latent using the prompt `a picture of a car'. Editing prompts denoted in blue. Arrows indicate the editing direction. The Stable Artist can act on explicit edits for local and global changes of the image, as well as abstract editing concepts. (Best viewed in color)}
      \label{fig:editing_concepts}
\end{figure*}
But back to $\mu$. It considers those dimensions of the prompt conditioned estimate that are relevant to the defined editing prompt $e$. To this end, $\mu$ scales the element-wise difference between the prompt conditioned estimate and edit conditioned estimate by $s_e$ for all elements where this difference is below or above a threshold $\lambda$ and equals $0$ otherwise: 
\begin{align}
&\mu(\mathbf{c}_p, \mathbf{c}_e; s_e, \lambda)= \nonumber \\
    &\begin{cases}
    \text{max}(1, |\phi|) ,& \text{where } \mathbf{\epsilon}_\theta(\mathbf{z}_t, \mathbf{c}_p) \ominus \mathbf{\epsilon}_\theta(\mathbf{z}_t, \mathbf{c}_e) \lessgtr \lambda \\
    0,              & \text{otherwise}
    \label{eq:threshold}
    \end{cases}\\ 
    &\text{with}\quad \phi = s_e 
    \big({\mathbf{\epsilon}_\theta(\mathbf{z}_t, \mathbf{c}_p) - \mathbf{\epsilon}_\theta(\mathbf{z}_t, \mathbf{c}_e)}\big)^{-1}
    \label{eq:safety_value}
\end{align}
with both larger absolute values of $\lambda$ and larger $s_e$ leading to a more substantial shift away from the prompt text. For positive guidance, values greater than the threshold are considered and vice versa for negative guidance. Consequently, the former should choose negative values for $\lambda$ and the latter positive ones.  

Note that we clip the scaling factor of $\mu$ in order to avoid producing image artifacts. Following~\cite{saharia2022photorealistic, ho2020denoising}, the values of each $\mathbf{x}$-prediction should adhere to the training bounds of $[-1,1]$ to prevent low fidelity images.

To offer even more control over the diffusion process, we make two adjustments to the methodology presented above. 
We add a warm-up parameter $\delta$ that will only apply  guidance $\gamma$ after an initial warm-up period in the diffusion process, i.e., $\gamma(\mathbf{z}_t, \mathbf{c}_p, \mathbf{c}_e) := \mathbf{0} \text{ if }t < \delta$. Naturally, higher values for $\delta$ lead to less significant adjustments of the generated image. 
If we aim to keep the overall composition of the image unchanged, selecting a sufficiently high $\delta$ ensures that only fine-grained details of the output are altered. 

Furthermore, we add a momentum term $\nu_t$ to the semantic guidance $\gamma$ in order to accelerate guidance over time steps for dimensions that are continuously guided in the same direction. Hence, $\gamma_t$ is defined as: 
\begin{align}
\gamma_t(\mathbf{z}_t, &\mathbf{c}_p, \mathbf{c}_e) = 
\mu(\mathbf{c}_p, \mathbf{c}_e; s_e, \lambda)  \psi(\mathbf{z}_t, \mathbf{c}_p, \mathbf{c}_e) + s_m \nu_t \label{eq:safety_guidance} 
\end{align}
with momentum scale $s_m \in [0,1]$ and $\nu$ being updated as 
\begin{equation}
\label{eq:safety_momentum}
    \nu_{t+1} = \beta_m  \nu_t + (1-\beta_m)  \gamma_t
\end{equation}
where $\nu_0 = \mathbf{0}$ and $\beta_m \in [0,1)$. Thus, larger $\beta_m$ lead to less volatile changes of the momentum. Momentum is already built up during the warm-up period, even though $\gamma_t$ is not applied during these steps.

\paragraph{Beyond One Direction.} Now we are ready to move beyond using just one direction towards multiple concepts $e_i$ and in turn combining multiple calculations of $\gamma_t$.

For all $e_i$, we calculate $\gamma_t^i$  as described above with each defining their own parameter values $\lambda^i$, $s_e^i$. The adjusted $\hat{\gamma}_t$ is the result of the weighted sum of all $\gamma_t^i$: 

\begin{equation}
     \hat{\gamma}_t(\mathbf{z}_t, \mathbf{c}_p; \mathbf{e}) = \sum\nolimits_{i \in I} g_i \gamma_t^i(\mathbf{z}_t, \mathbf{c}_p, \mathbf{c}_{e_i}) 
    \label{eq:mult_sumup}
\end{equation}

where $g_i$ sums up to one. In order to account for different warm-up periods $g_i$ is defined as $g_i\! =\! 0$ if $t < \delta_i$. However, momentum is built up using all edit prompt and applied once all warm-up periods are completed, i.e. $\forall \delta_i : \delta_i \geq t$. We provide a pseudo-code implementation of \textsc{Sega} in \cref{alg:sega}.  Please note that this notation makes the simplified assumption of one single warm-up period $\delta$ for all edit prompts $e_i$.


\subsection{Interactively Steering Semantics} 
Putting everything together, we start with an image generated from a prompt. Then, we iteratively perform adjustments with \textsc{Sega}, enforcing or suppressing concepts, thus steering the overall semantics in diffusion latent space. \cref{fig:arithmetics} provides a 2-dimensional, visual explanation of semantic guidance. Intuitively, we can understand the latent space as a composition of arbitrary sub-spaces representing semantic concepts. The unconditioned noise estimate (black dot) starts us off at some random point in the latent space without semantic grounding. The guidance corresponding to the prompt ``a portrait of a king'' represents a vector (blue vector) moving us into a portion of the latent space where the concepts `male' (human) and royal overlap, resulting in an image of a king. 
We can now further manipulate the generation process using \textsc{Sega}. From the unconditioned starting point, we get the directions of `male' and `female' (orange/green lines) using estimates conditioned on the respective prompts. If we subtract this inferred `male' direction from our prompt guidance and add the `female' one, we now reach a point in latent space at the intersection of the `royal' and `female' sub-spaces, i.e. a queen. This vector represents the final direction (red vector) resulting from semantic guidance. 

\begin{figure*}
\centering
     \includegraphics[width=.9\linewidth]{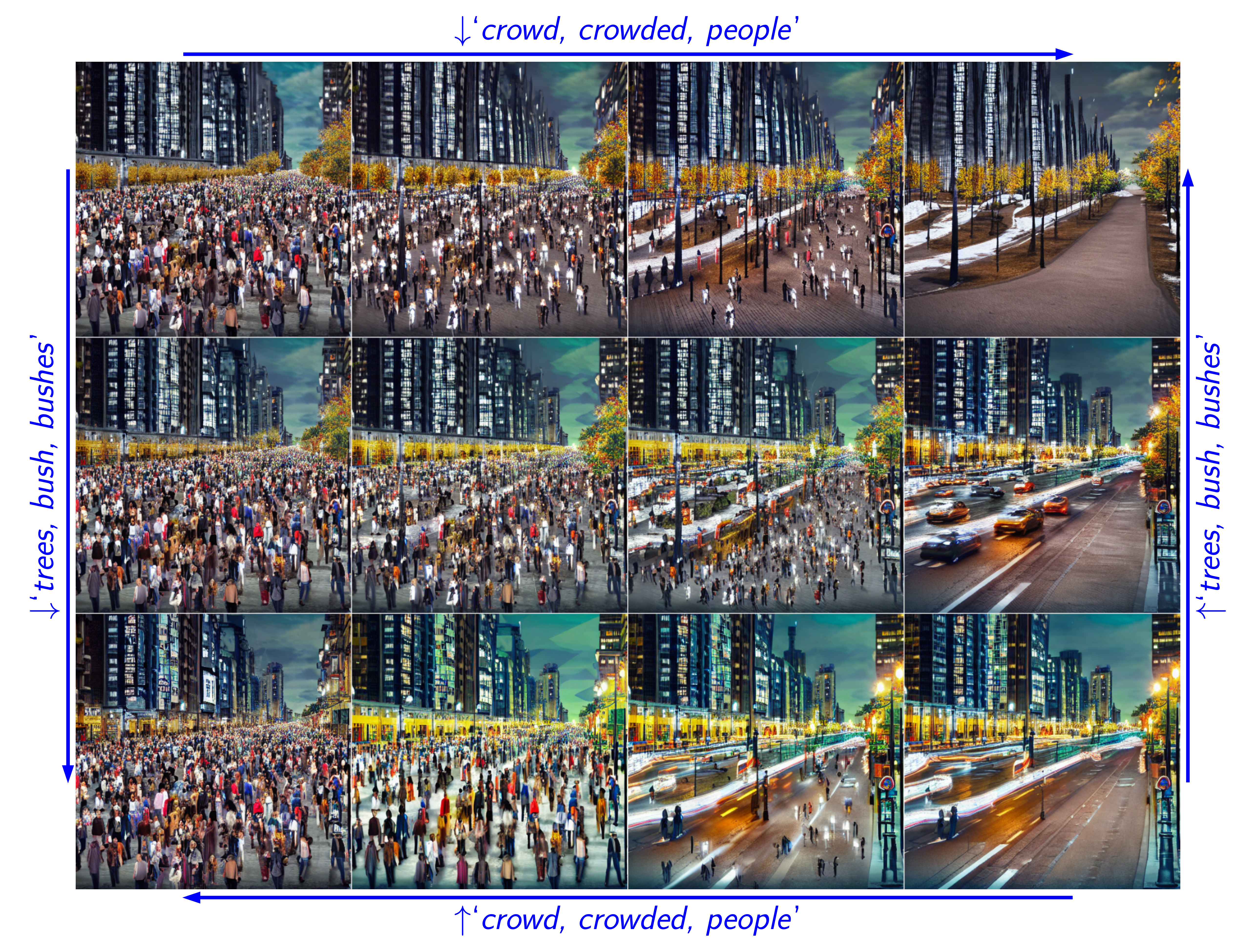}
     \caption{The Stable Artist offers strong control over the latent space and can gradually perform edits at the desired strength. All images generated from the same initial noise latent using the prompt `a crowded boulevard'. Editing prompts denoted in blue and are gradually increased in strength from left to right and top to bottom. (Best viewed in color)}
     \label{fig:progression}
\end{figure*}

In the next section, we now illustrate the Stable Artist. The illustrations are based on our own implementation of Stable Diffusion version~1.5 \footnote{\url{https://huggingface.co/runwayml/stable-diffusion-v1-5}}. The code, together with detailed demonstrations, is publicly available at \url{github.com/ml-research/semantic-image-editing}. 

\section{Interacting with Visual Concepts}
In contrast to existing work, the Stable Artist directly interacts with concepts implicitly learned by DMs. 
Consider e.g.~\cref{fig:arithmetics_example}. As one can see, the linear operations between noise estimates distilled using \textsc{Sega} 
are semantically grounded. Arithmetic relations between concepts like these have previously been observed for natural language embeddings. In the depicted example, editing an image of a king by guiding it away from the concept `\textit{male}' and towards the concept `\textit{female}' produces a compositionally similar image but replaces the king with a queen. This indicates that the latent space of diffusion models may inherently be disentangled to some extent, although further research in that direction is necessary to verify our assumption. 

Furthermore, the Stable Artist can simultaneously change multiple aspects of an image with minimal interference of different concepts, as shown in \cref{fig:examples} (top right) and \cref{fig:editing_concepts} (top mid). We discuss the subtle control offered by \textsc{Sega} in more detail in \cref{sec:control}. Additionally, \cref{fig:editing_concepts} highlights the versatile concepts that may be used to adjust images faithfully. These include concrete editing prompts that explicitly target certain portions of the image. In this regard, the stable artist supports both local edits to one of the depicted objects and global edits aimed at the environment of the image as a whole. Notably, multiple local and global edits may be combined arbitrarily.
On the other hand, the Stable Artist also helps uncover how the underlying DM ``interprets'' more complex concepts and gives further insight into learned representations. For example, adding the concept `\textit{carbon emissions}' to the generated car produces a seemingly much older vehicle with a presumably larger carbon footprint. 
Similarly, reducing `\textit{safety}' yields a convertible with no roof and likely increased horsepower.
Both these interpretations of the provided concepts with respect to cars are logically sound and provide valuable insights into the learned concepts of DMs. These experiments suggest a deeper natural language and image ``understanding'' that go beyond descriptive captions of images.

\section{Fine-grained Image Editing \& Composition}\label{sec:control}
One of our driving forces is the vision that
generative image creation and composition techniques should offer fine-grained control over the generated image. \textsc{Sega} targets the relevant portions of the image while not changing other aspects. Furthermore, the Stable Artist offers continuous control over the strength of the applied concept semantically reflected in the image. It is worth noting that we achieve this without masking the image, be it user-provided masks or those implicitly inferred from attention maps. Instead, we produce results of competitive quality by targeting the relevant latent dimensions alone. 

This level of control can be seen throughout the images shown in the present work. Considering the top right example in \cref{fig:examples}, we make two local changes to the image composition and a global style change affecting the entire image. All three operations are performed simultaneously in the same latent space, and nonetheless, the majority of the image remains unchanged down to small details like the structure of clouds, the bush in the foreground, the castle, or the hill it stands on. We can observe a similar level of control in \cref{fig:editing_concepts}, where the fore- and background remain largely unaffected by changes to the car and vice versa. 

We investigated this further in \cref{fig:progression}, together with possible linear continuity of image alterations.  Starting from the original image in the top left corner, we remove one concept from the image following the x- or y-axis. In both cases, the Stable Artist continually reduces the number of people or trees until the respective concept is removed entirely. Again, the rest of the image remains fairly consistent, especially the concept that is not targeted. Going even further, a similar level of control is also possible with an arbitrary mixture of applied concepts. Let us consider the second row of \cref{fig:progression}, for example. The number of trees is kept at a medium level in that the row of trees on the left side of each image is always removed, but the larger tree on the right remains. While keeping the number of trees stable, we can still gradually remove the crowd at the same time. 

\begin{figure*}
\centering
     \includegraphics[width=.9\linewidth]{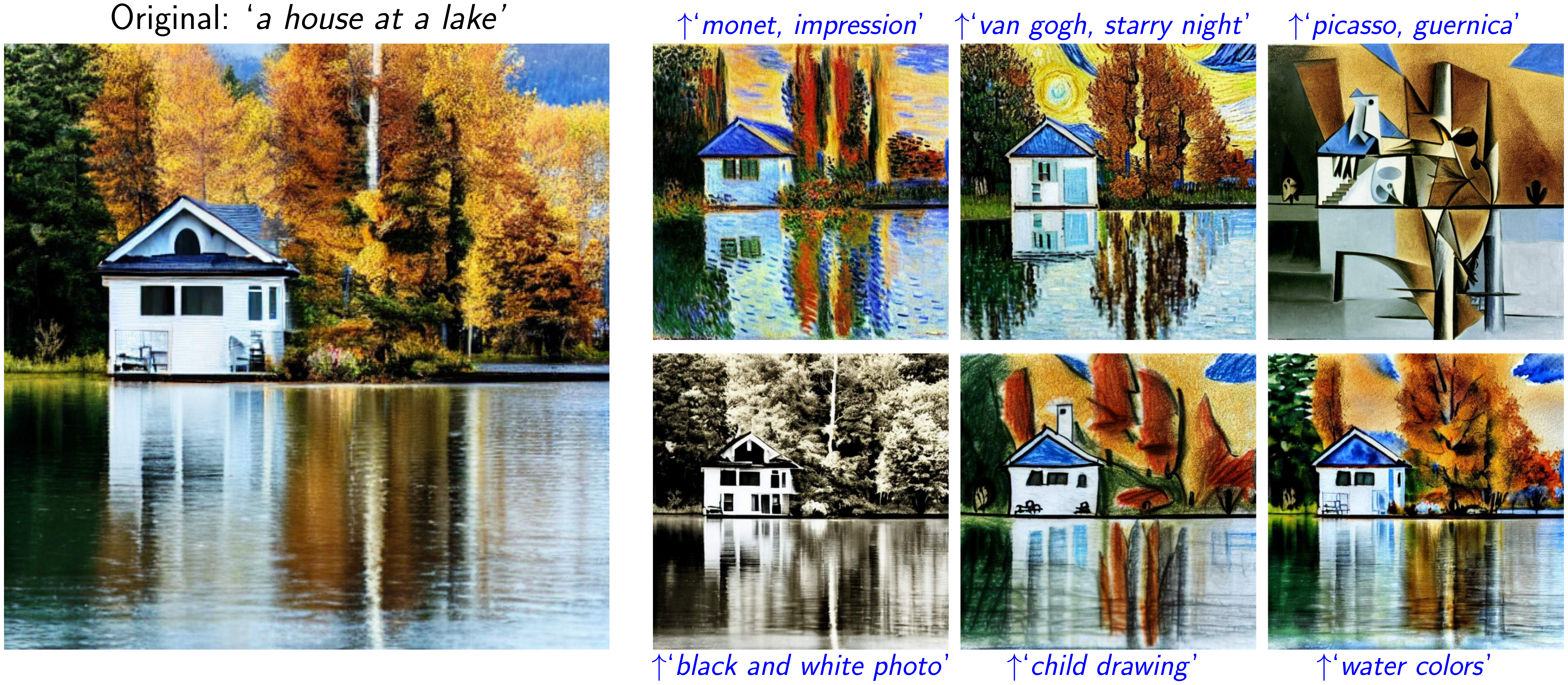}
     \caption{Style transfer performed by the Stable Artist. All images generated from the same initial noise latent using the prompt `a house at a lake'. Editing prompts denoted in blue. Arrows indicate the editing direction. (Best viewed in color)}
     \label{fig:style_transfer}
\end{figure*}
\begin{figure*}
\centering
     \includegraphics[width=.85\linewidth]{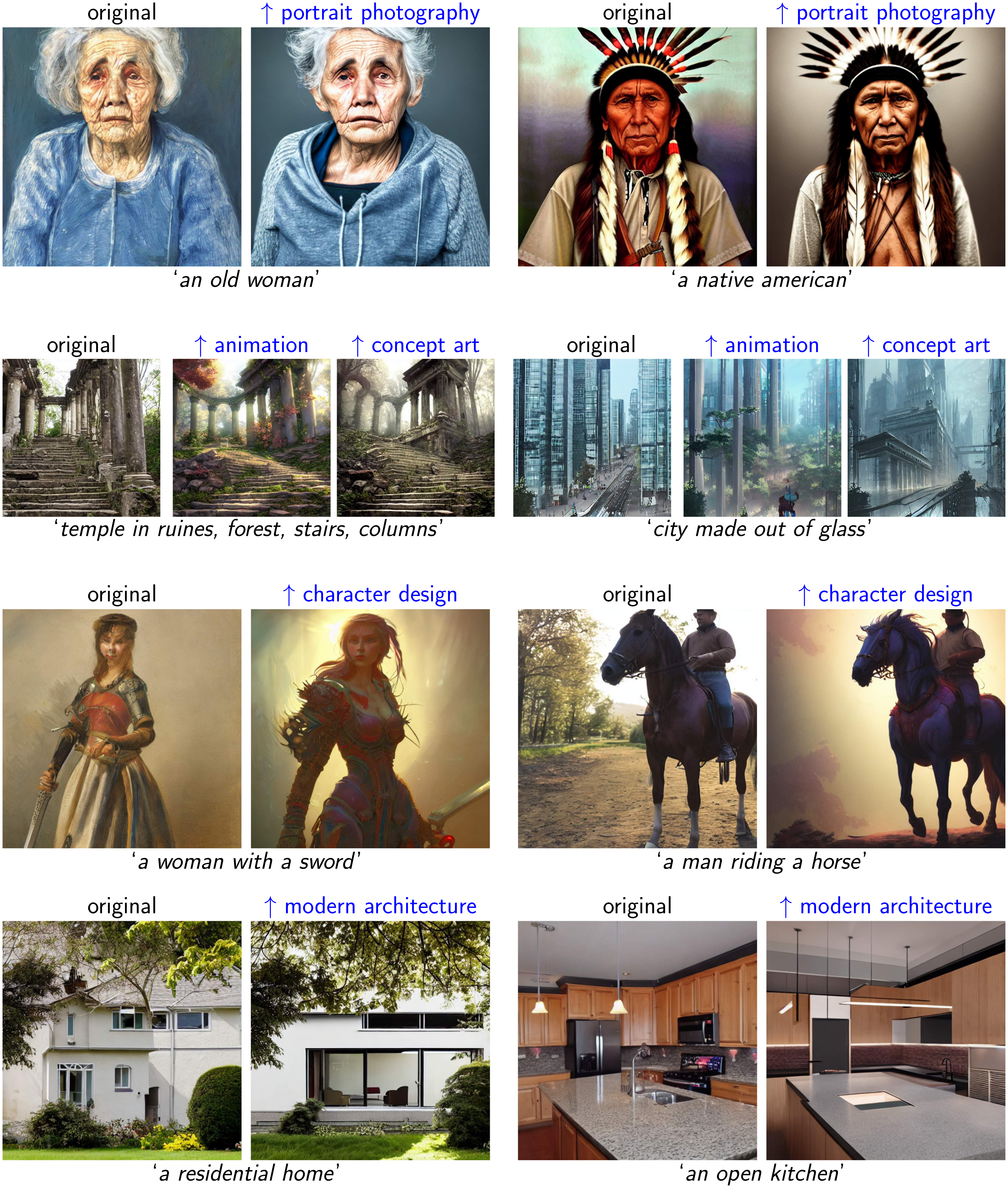}
     \caption{Latent style and image quality optimization using the Stable Artist. All image pairs/triplets are generated from the same initial noise latent and the text prompt under the images. Images are guided towards the average prompt embeddings of high-quality images belonging to the category stated in blue. (Best viewed in color)}
     \label{fig:latent_image_optimization}
\end{figure*}
\begin{figure*}
    \centering
    \includegraphics[width=0.95\linewidth]{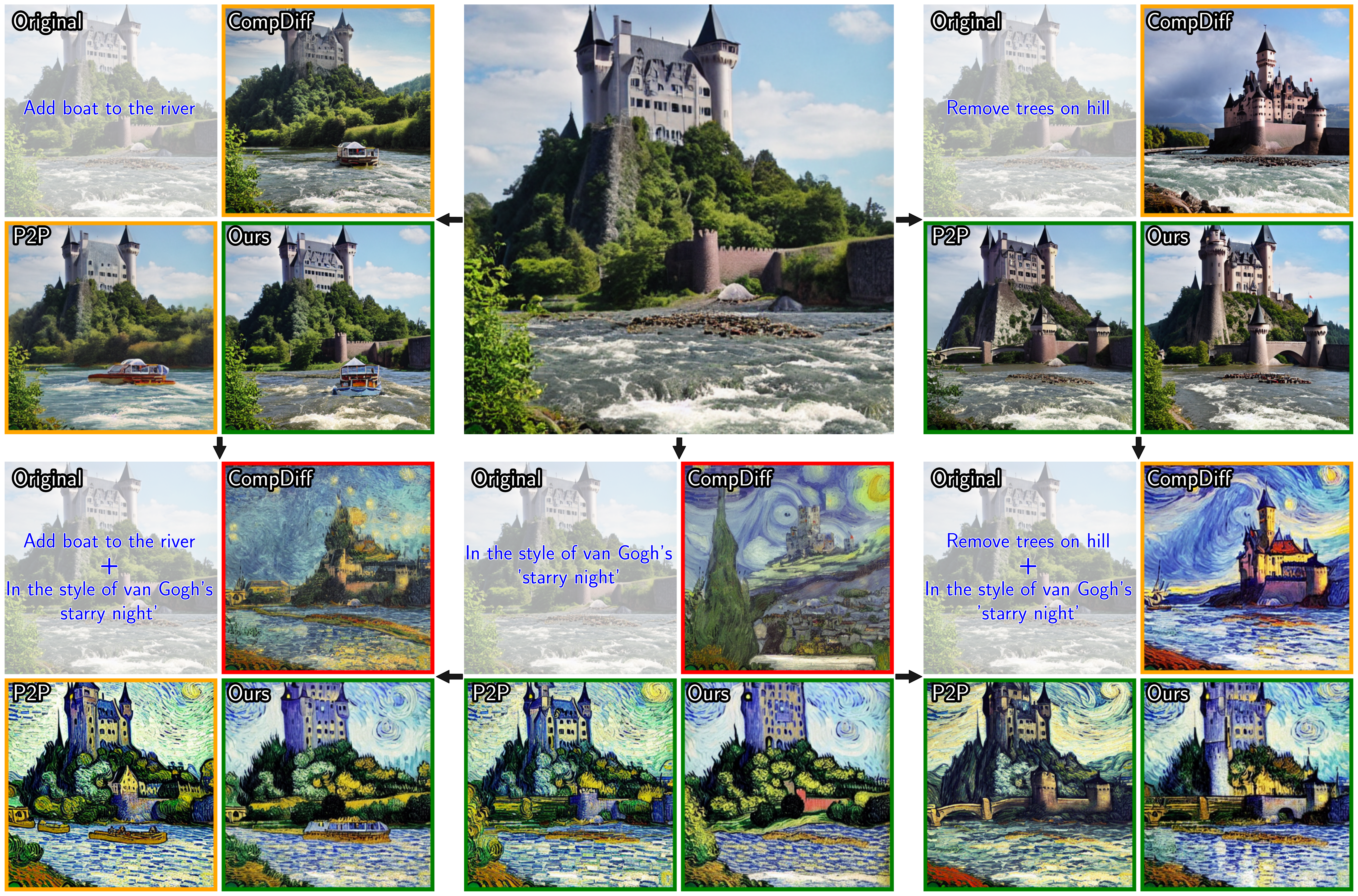}
    \caption{Comparison of image editing between Prompt-to-Prompt (P2P) \cite{hertz2022prompt}, Composable Diffusion (CompDiff)\cite{liu2022Compositional}, and the Stable Artist with \textsc{Sega} (ours). All images were generated from the same initial noise latent using the prompt `a castle next to a river'. Colored boxes around the image denote the quality of the edit with respect to the editing task and the goal of minimal changes to the unaffected parts of the image. (Best viewed in color)}
    \label{fig:comparisions}
\end{figure*}

\section{Latent Image Optimization}

\textsc{Sega} can be used to influence the overall style of an image as well as its quality with respect to certain artistic conceptions. \cref{fig:style_transfer} demonstrates examples of style transfer in which a photorealistic image is adjusted to reflect a different style of photography or painting. The Stable Artist faithfully reproduces the styles of well-known artists, as well as more abstract instructions. Again, the overall image composition remains largely unchanged, enabling a real transfer of the original image to the target style. This sets \textsc{Sega} clearly apart from more simplistic techniques that only append the style instruction to the original prompt. 

With Stable Artist, we can go even further than vanilla style transfer and actually optimize the overall look and quality of the generated output for arbitrary types of images directly in latent space. Creating high-fidelity images often requires a large amount of error-prone prompt engineering for every specific type of image and, in turn, results in long and obfuscated prompts. Instead, we can use a short and concise prompt and directly optimize the style through semantic guidance. 
To that end, we collected prompts known to produce high-quality results\footnote{Prompts taken from \url{https://mpost.io/best-100-stable-diffusion-prompts-the-most-beautiful-ai-text-to-image-prompts/}} for five different types of images in \textit{portrait photography, animation, concept art, character design}, and \textit{modern architecture}. We calculated the text embeddings for a set of prompts and took the average embedding per category as guidance conditioning. Exemplary results are depicted in \cref{fig:latent_image_optimization}. The results are of high quality and stay close to the original image, but accurately reflect the targeted artistic direction. 

\section{Qualitative Comparison}
To investigate whether the Stable Artist provides more control to the artist, we compared it qualitatively to related techniques for text-to-image diffusion. While the fine-grained control and probing of the latent diffusion space are unique to \textsc{Sega}, 
closest to the Stable Artist are Composable Diffusion \cite{liu2022Compositional} and Prompt-to-Prompt \cite{hertz2022prompt}.
To compare them, we assess the performance of each approach on five different tasks depicted in \cref{fig:comparisions}. These consist of additive and subtractive image composition, style transfer, and the combination of composition and style changes.

All three methods perform well in adding elements to the image (\cref{fig:comparisions} top left). Nonetheless, the Stable Artist is the only one keeping the riverside brick stone wall from the original image and also makes the least visible alterations to the castle and river. In the case of removing elements (\cref{fig:comparisions} top right), Composable Diffusion alters large portions of the image composition, whereas P2P and the Stable Artist stay more faithful to the original image. This experiment further supports an interesting observation of the Stable Diffusion latent space. The model tends to prefer covering up image components targeted for removal with new details instead of generating the material that would be revealed behind or beneath the removed contents. 

On the style transfer task (\cref{fig:comparisions} bottom mid), Composable Diffusion is incapable of combining the original image composition with the target style. For progressively stronger conditioning on van Gogh's starry night, the approach simply shifts towards replicating the original painting and loses the image composition of the castle. However, P2P and the Stable Artist both create a faithful fusion between the content's starting image and the target style. 

When combining editing and style transfer (\cref{fig:comparisions} bottom left \& right), Composable Diffusion may generate images that do not fulfill any of the input descriptions. \textsc{Sega} and P2P produce satisfactory results; however, the latter is limited to one editing prompt. Consequently, we generated these results by supplying both editing tasks in the same prompt for P2P. This also results in artifacts like multiple boats being added, although only one was specified. 

Overall, Stable Artist outperforms Composable Diffusion in general image editing and composition on a variety of tasks. We achieve better or at least equally good performance in targeted editing as P2P but natively support conditioning on multiple concepts. Furthermore, \textsc{Sega} does not rely on token-based attention and is therefore applicable to any conditioning beyond natural language, such as conditioning on images, averaged textual embeddings or embeddings calculated through textual inversion. We discuss this further in \cref{sec:conclusion}. 

\section{Broader Impact on Society}
Recent developments in text-to-image models \cite{ramesh2022hierarchical, nichol2022glide, saharia2022photorealistic} have the potential for far-reaching impact on society, both positive and negative, when deployed in applications such as image generation, image editing, or search engines.
Previous research \cite{bianchi2022large, schramowski2022safe} described many potential negative societal implications that may arise due to the careless use of such large-scale generative models. Many of these problems can be attributed to the noisy, large-scale datasets these models rely on.
Since recent text-to-image models, such as stable diffusion, are trained on web-crawled data containing inappropriate content \cite{schuhmann2022laion}, they are no exception to this issue. Specifically, current versions of stable diffusion show signs of inappropriate degeneration \cite{schramowski2022safe}. While Schramowski~\textit{et al.}~utilize the model's notion of inappropriateness to steer the model away from generating related content, it is noteworthy that we introduce an approach that could also be used to guide image generation toward inappropriate material. However, on the positive side, \textsc{Sega} could also have the potential to increase e.g. fairness, by detecting and steering certain related concepts. Therefore, we advocate for further research in this direction.

Another frequently voiced point of criticism, is the notion that generative models like stable diffusion are replacing human artists and illustrators. At first glance, the great results produced by these models might warrant this impression. Examples like the DALL-E generated Cosmopolitan cover\footnote{\url{https://www.cosmopolitan.com/lifestyle/a40314356/dall-e-2-artificial-intelligence-cover/}} subtitled with the phrase ``And it only took 20 seconds to make'' certainly seem to support this point.
However, looking more closely at this example reveals that creating this cover still involved multiple hours spent by a human user interacting with the model. Consequently, we argue that generative models remain a tool to be used by humans for creating artwork.
The generative process as a whole still requires a substantial amount of iterative human feedback and creative thinking. The introduced Stable Artist increases the interaction capabilities in these processes.

\section{Conclusions}\label{sec:conclusion}
We presented the Stable Artist for directly interacting with concepts in DM's latent space. To this end, we introduced semantic guidance (SEGA) that allows one to influence/steer the diffusion process along several directions. We demonstrated that the Stable Artist using \textsc{Sega} offers fine-grained control over the generated image for performing sophisticated image composition and editing. 

The Stable Artist covers several exciting avenues for future work. 
For instance, one should investigate more closely how concepts are represented in the latent space of DM's and how to target and quantify them. 
More importantly, automatically detecting concepts could provide novel insights and toolsets to mitigate biases, as well as enacting privacy concerns of real people memorized by the model.

\paragraph{Acknowledgments.}
This research has benefited from the Hessian Ministry of Higher Education, Research, Science and the Arts (HMWK) cluster projects ``The Third Wave of AI'' and hessian.AI, from the German Center for Artificial Intelligence (DFKI) project ``SAINT'', the Federal Ministry of Education and Research (BMBF) project KISTRA (reference no. 13N15343), as well as from the joint ATHENE project of the HMWK and the BMBF ``AVSV''.

{\small
\bibliographystyle{ieee_fullname}
\bibliography{bibliography}

\begin{thebibliography}{10}\itemsep=-1pt

\bibitem{agirre-etal-2012-semeval}
Eneko Agirre, Daniel Cer, Mona Diab, and Aitor Gonzalez-Agirre.
\newblock {S}em{E}val-2012 task 6: A pilot on semantic textual similarity.
\newblock In {\em *{SEM} 2012: The First Joint Conference on Lexical and
  Computational Semantics {--} Volume 1: Proceedings of the main conference and
  the shared task, and Volume 2: Proceedings of the Sixth International
  Workshop on Semantic Evaluation ({S}em{E}val 2012)}, pages 385--393, 2012.

\bibitem{bianchi2022large}
Federico Bianchi, Pratyusha Kalluri, Esin Durmus, Faisal Ladhak, Myra Cheng,
  Debora Nozza, Tatsunori Hashimoto, Dan Jurafsky, James Zou, and Aylin
  Caliskan.
\newblock Easily accessible text-to-image generation amplifies demographic
  stereotypes at large scale.
\newblock {\em CoRR}, abs/2211.03759, 2022.

\bibitem{hertz2022prompt}
Amir Hertz, Ron Mokady, Jay Tenenbaum, Kfir Aberman, Yael Pritch, and Daniel
  Cohen{-}Or.
\newblock Prompt-to-prompt image editing with cross attention control.
\newblock Preprint at \url{https://arxiv.org/abs/2208.01626}, 2022.

\bibitem{ho2020denoising}
Jonathan Ho, Ajay Jain, and Pieter Abbeel.
\newblock Denoising diffusion probabilistic models.
\newblock In Hugo Larochelle, Marc'Aurelio Ranzato, Raia Hadsell,
  Maria{-}Florina Balcan, and Hsuan{-}Tien Lin, editors, {\em Proceedings of
  the Advances in Neural Information Processing Systems: Annual Conference on
  Neural Information Processing Systems ({NeurIPS})}, 2020.

\bibitem{ho2022classifier}
Jonathan Ho and Tim Salimans.
\newblock Classifier-free diffusion guidance.
\newblock {\em CoRR}, abs/2207.12598, 2022.

\bibitem{levy-goldberg-2014-linguistic}
Omer Levy and Yoav Goldberg.
\newblock Linguistic regularities in sparse and explicit word representations.
\newblock In {\em Proceedings of the Eighteenth Conference on Computational
  Natural Language Learning}, pages 171--180, 2014.

\bibitem{liu2022Compositional}
Nan Liu, Shuang Li, Yilun Du, Antonio Torralba, and Joshua~B. Tenenbaum.
\newblock Compositional visual generation with composable diffusion models.
\newblock In {\em Proceedings of European Conference on Computer Vision
  ({ECCV})}, 2022.

\bibitem{mikolov-etal-2013-linguistic}
Tomas Mikolov, Wen-tau Yih, and Geoffrey Zweig.
\newblock Linguistic regularities in continuous space word representations.
\newblock In {\em Proceedings of the 2013 Conference of the North {A}merican
  Chapter of the Association for Computational Linguistics: Human Language
  Technologies}, pages 746--751, 2013.

\bibitem{nichol2022glide}
Alexander~Quinn Nichol, Prafulla Dhariwal, Aditya Ramesh, Pranav Shyam, Pamela
  Mishkin, Bob McGrew, Ilya Sutskever, and Mark Chen.
\newblock {GLIDE:} towards photorealistic image generation and editing with
  text-guided diffusion models.
\newblock In {\em Proceedings of the International Conference on Machine
  Learning ({ICML})}. {PMLR}, 2022.

\bibitem{ramesh2022hierarchical}
Aditya Ramesh, Prafulla Dhariwal, Alex Nichol, Casey Chu, and Mark Chen.
\newblock Hierarchical text-conditional image generation with {CLIP} latents.
\newblock Preprint at \url{https://arxiv.org/abs/2204.06125}, 2022.

\bibitem{rombach2022High}
Robin Rombach, Andreas Blattmann, Dominik Lorenz, Patrick Esser, and
  Bj{\"{o}}rn Ommer.
\newblock High-resolution image synthesis with latent diffusion models.
\newblock In {\em Proceedings of the {IEEE/CVF} Conference on Computer Vision
  and Pattern Recognition ({CVPR})}, 2022.

\bibitem{saharia2022photorealistic}
Chitwan Saharia, William Chan, Saurabh Saxena, Lala Li, Jay Whang, Emily
  Denton, Seyed Kamyar~Seyed Ghasemipour, Burcu~Karagol Ayan, S.~Sara Mahdavi,
  Rapha~Gontijo Lopes, Tim Salimans, Jonathan Ho, David~J. Fleet, and Mohammad
  Norouzi.
\newblock Photorealistic text-to-image diffusion models with deep language
  understanding.
\newblock {\em CoRR}, abs/2205.11487, 2022.

\bibitem{schramowski2022safe}
Patrick Schramowski, Manuel Brack, Björn Deiseroth, and Kristian Kersting.
\newblock Safe latent diffusion: Mitigating inappropriate degeneration in
  diffusion models.
\newblock {\em arXiv preprint arXiv:2211.05105}, 2022.

\bibitem{schuhmann2022laion}
Christoph Schuhmann, Romain Beaumont, Richard Vencu, Cade~W Gordon, Ross
  Wightman, Theo Coombes, Aarush Katta, Clayton Mullis, Mitchell Wortsman,
  Patrick Schramowski, Srivatsa~R Kundurthy, Katherine Crowson, Ludwig Schmidt,
  Robert Kaczmarczyk, and Jenia Jitsev.
\newblock Laion-5b: An open large-scale dataset for training next generation
  image-text models.
\newblock In {\em Thirty-sixth Conference on Neural Information Processing
  Systems Datasets and Benchmarks Track}, 2022.

\end{thebibliography}
}
\clearpage

\end{document}